\documentclass[10pt,twocolumn,letterpaper]{article}
\usepackage{wacv}
\usepackage{times}
\usepackage{epsfig}
\usepackage{graphicx}
\usepackage{amsmath}
\usepackage{amssymb}
\usepackage{amsthm}
\usepackage{booktabs}
\usepackage{algorithm}
\usepackage[noend]{algpseudocode}
\usepackage{bm}
\usepackage{subfig}

\theoremstyle{definition}

\newcommand{\argmax}{\mathop{\mathrm{argmax}}\limits}
\def\wacvPaperID{12}

\wacvalgorithmstrack   

\wacvfinalcopy 

\ifwacvfinal
\usepackage[breaklinks=true,bookmarks=false]{hyperref}
\else
\usepackage[pagebackref=true,breaklinks=true,colorlinks,bookmarks=false]{hyperref}
\fi

\begin{document}

\title{MetaMax: Improved Open-Set Deep Neural Networks via Weibull Calibration}

\author{Zongyao Lyu, Nolan B. Gutierrez, and William J. Beksi\\
The University of Texas at Arlington\\
Arlington, TX, USA\\
{\tt\small zongyao.lyu@mavs.uta.edu, nolan.gutierrez@mavs.uta.edu, william.beksi@uta.edu}
}

\maketitle
\thispagestyle{empty}

\begin{abstract}
Open-set recognition refers to the problem in which classes that were not seen
during training appear at inference time. This requires the ability to identify
instances of novel classes while maintaining discriminative capability for
closed-set classification. OpenMax was the first deep neural network-based
approach to address open-set recognition by calibrating the predictive scores
of a standard closed-set classification network.  In this paper we present
MetaMax, a more effective post-processing technique that improves upon
contemporary methods by directly modeling class activation vectors. MetaMax
removes the need for computing class mean activation vectors (MAVs) and
distances between a query image and a class MAV as required in OpenMax.
Experimental results show that MetaMax outperforms OpenMax and is comparable in
performance to other state-of-the-art approaches.
\end{abstract}

\section{Introduction}
\label{sec:introduction}
Image classification with deep neural networks has made significant progress
\cite{krizhevsky2012imagenet,simonyan2014very,he2016deep,huang2017densely}.
However, the majority of the work is based on a closed-set assumption where
training datasets are expected to include all the classes that may be
encountered in the environments in which the vision system will be deployed.
Yet, this assumption cannot be guaranteed in real-world environments where
samples from unknown classes not seen during training may appear during testing
and cause system failure
\cite{pinto2008real,torralba2011unbiased,zhang2014predicting}.

\begin{figure}
\centering
\setlength{\abovecaptionskip}{0.08cm}
\includegraphics[width=\columnwidth]{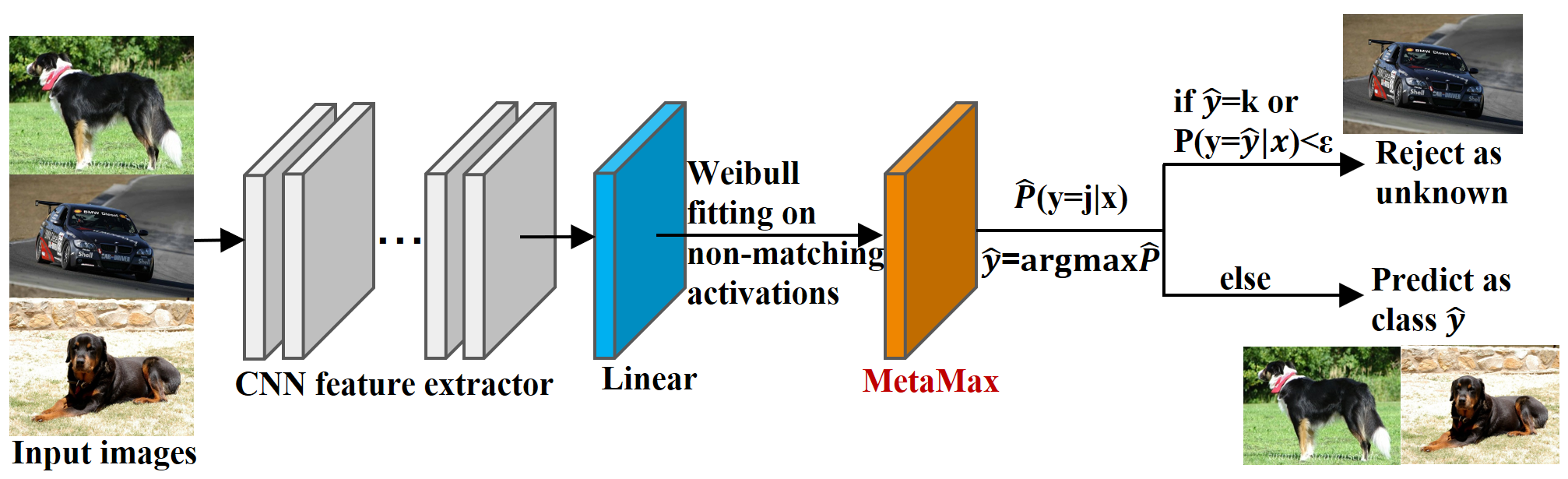}
\caption{MetaMax utilizes extreme value theory-based meta-recognition for
post-recognition analysis to discover the presence of unknown classes while
preserving closed-set classification accuracy.}
\label{fig:metamax_overview}
\end{figure}

To address the limitation of closed-set classification, open-set recognition
(OSR) has been introduced \cite{scheirer2012toward}. OSR describes the scenario
where incomplete knowledge of the world is present during training, and new
classes can appear during testing. Not only does this require the model to
maintain the capability of accurately classifying known classes, but it must
also be able to effectively identify unknown classes.  OSR has recently gained
significant attention in the research community
\cite{geng2020recent,jain2014multi,scheirer2014probability,bendale2015towards,
bendale2016towards,dietterich2017steps,amodei2016concrete}.

With the introduction of the formal definition of OSR, Scheirer et al.
\cite{scheirer2012toward} presented a ``1-vs-Set Machine" which extends the
1-class and binary SVM in a way that better supports OSR. Bendale and Boult
\cite{bendale2016towards} proposed the first deep learning-based solution for
OSR by introducing OpenMax as an alternative to SoftMax. OpenMax can be
applied directly on a standard closed-set classification network to calibrate
its output score vectors, thus enabling a closed-set classifier to perform OSR.
It does not require accessing additional training data, updating the training
procedure, or modification of the network architecture. 

Inspired by OpenMax, we introduce MetaMax, a method that utilizes extreme value
theory-based meta-recognition \cite{scheirer2011meta,scheirer2012learning} as a
post-recognition analysis technique to assist the identification of unknown
classes, Fig.~\ref{fig:metamax_overview}. More specifically, we directly model
non-match scores from the network outputs as opposed to OpenMax's method of
explicitly calculating and storing mean activation vectors for each class. By
modeling the non-match distributions, we can assemble additional data points for
building a Weibull model and therefore achieve more accurate parameters for each
constructed model.

To the best of our knowledge, MetaMax is the \textit{only alternative} to OpenMax.
Different from most work in this area that either modifies the network architecture
or requires auxiliary training data that can be hard to acquire in practice,
we present a calibration tool that can be \textit{readily plugged into various
classification networks thus enabling them to perform OSR without additional
overhead}.

\section{Related Work}
\label{sec:related_work-ch03}

\subsection{Open-Set Recognition}
Open-set recognition is the extension of object recognition from traditional
closed-set conditions to open-set conditions where classes outside of the
training set can appear during inference. Many approaches have been proposed
to address the issue of OSR \cite{geng2020recent}. Early attempts mainly
focused on adapting traditional closed-set machine learning classifiers. With
the introduction of the formal definition of open-set recognition, Sheirer
et al. \cite{scheirer2012toward} presented a ``1-vs-Set Machine" which extends
the 1-class and binary support vector machine (SVM) in a way that better
supports OSR. In proceeding work, Sheirer et al. \cite{scheirer2014probability}
introduced a novel compact abating probability (CAP) model where the probability
of class membership abates as data points move away from known data space toward
the unknown open space. They further developed this idea by combining CAP with
statistical extreme value theory for probability estimates and score calibration.
The resulting method, Weibull-calibrated SVM, provides a probability score for
rejecting unknown observations and achieves much better performance on multiclass
OSR.

Bendale and Boult \cite{bendale2015towards} expanded upon open-set concepts
by presenting the problem of open-world recognition. In addition to formally
defining the issue of open-world recognition, Bendale and Boult showcased a
class-incremental learner in an open-world model. Finally, the authors proved
that weighted sums of compact abating probability models can have an arbitrarily
small ``open-space risk." However, the nearest neighbor approach presented in
their work is prone to over-fitting. Bendale and Boult \cite{bendale2016towards}
proposed the first solution for OSR with deep neural networks by introducing
OpenMax as an alternative to SoftMax. OpenMax is simple and effective, and
does not require changing the classification network architecture. Our proposed
MetaMax is a direct extension and improvement of OpenMax.

Other works have proposed incorporating techniques from various domains to
tackle OSR. For example, synthetic samples were first utilized in a training
set as unknown samples by Ge et al. \cite{ge2017generative}.
Among many generative approaches (e.g., \cite{ditria2020opengan,kong2021opengan}),
Oza and Patel \cite{oza2019c2ae} were the first to introduce class-conditioned
autoencoders. Sun et al. \cite{sun2020conditional} improved on these results
by training an end-to-end model thus eliminating the data preprocessing stages.
Perera et al. \cite{perera2020generative} showed that another benefit of
synthetic samples is that they can be used to augment data during inference.
Zhang et al. \cite{zhang2020hybrid} proposed an OpenHybrid framework
to learn a joint representation for discriminative and generative components,
Cao et al. \cite{cao2021open} directly extended upon Gaussian mixture variational
autoencoders \cite{dilokthanakul2016deep} by employing image classification
under open-set conditions. Zhou et al. \cite{zhou2021learning} propose to learn
data placeholders to anticipate open-set class data, thus transforming closed-set
training into open-set training.

\subsection{Anomaly Detection}
Anomaly detection (aka outlier detection) aims to find rare observations which
differ significantly in pattern from the majority of data
\cite{chandola2009anomaly,chalapathy2019deep}. Autoencoders
\cite{kingma2013auto} or adversarial autoencoders
\cite{makhzani2015adversarial} are widely used in deep neural network-based
anomaly detection algorithms by means of their reconstruction errors
\cite{zhou2017anomaly,pidhorskyi2018generative}. In addition, they are utilized
with Gaussian mixture models by combining the induced reconstruction error and
learned latent representation for unsupervised anomaly detection
\cite{zong2018deep}. Rather than discriminating between known classes, anomaly
detection typically focuses on the identification of unknowns. Thus, it is
possible to incorporate methods for anomaly detection into open-set recognition
as a component that addresses the part of identifying unknown classes.


\subsection{Learning With Reject Option}
Learning with reject option \cite{chow1970optimum,bartlett2008classification,
herbei2006classification,geifman2017selective} still works under closed-set
assumption, with the goal of rejecting an input sample due to its low
confidence thus avoiding categorizing a sample of one class as a member of
other classes. Therefore, this task should not be confused with the OSR problem
we study here.

\subsection{Calibration}
Calibration refers to the problem of obtaining a model that has a predictive
probability that reflects the true correctness likelihood. Conventional
calibration methods \cite{naeini2015obtaining,guo2017calibration} are all under
closed-set assumptions, therefore they are not addressing the same problem as
studied in this work. A recent work \cite{lyu2022evaluating} has also shown
that traditional calibration methods perform poorly under open-set conditions.
Thus, we do not consider conventional calibration methods for comparison in
this work.

\section{Preliminaries}
\label{sec:preliminaries}

\subsection{OpenMax}
\label{subsec:OpenMax}
OpenMax introduced the concept of \textit{activation vectors}, which are the
output values from the penultimate layer of a neural network.
Concretely, let $\bm{v}(x) = v_1(x),\ldots,v_K(x)$ be the level of activation
for each of the $K$ classes seen during training and let the unknown classes
begin at index 0. OpenMax obtains the activation vectors by saving the output of
$C_{\bm{\theta}}$'s layer immediately before the SoftMax layer. It then
separates the activation vectors into $K$ clusters representing $K$ known
classes. For each of the clusters, the distance between every activation vector
and the cluster mean (i.e., \textit{class mean activation vector}) is
calculated.

OpenMax proceeds to compute $K$ Weibull models by fitting each model to the
largest distances of the correctly classified samples. During inference, the
$K$ Weibull models are used to revise the top few highest activations to obtain
a new vector $\hat{\bm{v}}(x)$. OpenMax then defines an additional activation
for class $K+1$ thereby producing a score vector for all $K+1$ classes. The
class corresponding to the highest probability is the prediction of the OpenMax
algorithm. If the predicted class label for an input is indexed at 0, then it
is considered to be an unknown class.

\section{MetaMax}
\label{sec:metamax}
MetaMax is derived from contributions in class descriptors and the modeling of
non-match distributions for the purpose of calibrating OSR.   

\subsection{Class Representations}
\label{subsec:class_representations}
Object recognition can be mapped to the problem of determining \emph{match}
scores between an input image and a \textbf{class descriptor} such as the class
mean activation vector (MAV) used in OpenMax. Given an activation vector for a
correctly classified image by a closed-set classifier, we consider the largest
score as the \textbf{match} score and the rest of the scores in the vector as the
\textbf{non-match} scores. The match and non-match scores from all the training
data constitute the \textbf{match distribution} and \textbf{non-match distribution},
respectively.

\subsection{Modeling of Rare Events}
\label{subsec:modeling_of_rare_events}
Given a particular input, OpenMax finds the distance between each sample and
its respective MAV. The farthest, i.e., the rarest, distances are used to build
per class Weibull models. These models are then utilized to compute revised
activation vectors that OpenMax uses to provide probability estimates for the
unknown rejection. Yet, if data is limited these largest distances may
misrepresent the data manifold, produce incorrect activation vectors, and thus
hinder the Weibull model's accuracy.

When building the Weibull model for each class, OpenMax finds a single data
point by modeling the largest distance. We eliminate this limitation by
modeling the non-match distribution to provide multiple data points per input.
Concretely, we model the non-match distribution via non-match scores to obtain
$K-1$ data points per vector.

The first step of MetaMax is to obtain a Weibull model fitted for each class as
shown in Algorithm~\ref{alg:meta_max_algorithm}. To do this, we first train a
regular classification network.  Next, we collect all the activation vectors
and separate them into $K$ sets, one for each known class.  Each of these sets
have shape $N_j$ by $K$, where $N_j$ is the number of samples belonging to
class $j$. We proceed to remove column $j$ for each set, where $j$ indicates
the index of the match score (line 1 of
Algorithm~\ref{alg:meta_max_algorithm}).

The next step is to concatenate all of the non-match scores to obtain $\bm{W}$
resulting in $N \cdot (K-1)$ data points (line 2). We employ the FitHigh
function of LibMR \cite{libmr2018} to find the parameters of the Weibull
distribution using only the $q$ highest activations among the non-match class
activations. LibMR is an open-scource library for Meta-Recognition, and FitHigh
is the function in LibMR for building Weibull models. Lastly, we pass $\bm{W}$
to FitHigh to obtain the model $p_j$ for class $j$ (line 3).

\begin{algorithm}
\begin{algorithmic}[1]
\Require{LibMR FitHigh function}
\Require{Activation set $\bm{A} \in \mathcal{R}^{N \times K} $ for inputs $\bm{X}
\in \mathcal{R}^{H \times W \times c}$ belonging to class $j$}
\Require{$q$, the number of highest activations to model}
\Ensure{Weibull model for class $j$}
\State{{Obtain only non-match scores by removing column $j$ from activation set
$\bm{A}$ to acquire new activation vectors $\tilde{\bm{A}}
\in \mathcal{R}^{N \times (K - 1)}$}}
\State{Concatenate all non-match scores to obtain $N \cdot (K-1)$ data points
resulting in $\bm{W}$}
\State{Weibull Fit $p_j = (\rho_j,\kappa_j,\lambda_j)$ = FitHigh$(\bm{W}, q)$}
\end{algorithmic}
\caption{Weibull calibration for class activations.}
\label{alg:meta_max_algorithm}
\end{algorithm}

We derived a working algorithm by analyzing how non-match activations affect
calibration in MetaMax's inference step. MetaMax's inference step
(Algorithm~\ref{alg:meta_estimation}) uses the per-class Weibull models to make
a decision on whether to accept the network's SoftMax scores or to reject the
input. In lines 3 through 6, we modify the modulation vector $\bm{m}$, which has
been initialized to all ones. We obtain the key second step of MetaMax by
extending OpenMax to allow for the modeling of non-match activations. The main 
difference concerns the modulation vector $\bm{m}$'s construction, where the
Weibull cumulative distribution function is applied to modify the activation
vector. Since we do not model the highest class activation, we effectively
remove the influence of the highest activation on the unknown activation $a_0$
by skipping the index associated with the highest class activation. The
consequence of this modification is that the unknown class activation (defined
in line 8) simply adds zero when index $i$ is equal to the index of the highest
class activation. The new activations are then passed to the standard SoftMax
function to produce the class probabilities (line 9). Finally, a prediction is
found and a decision to accept or reject is made (line 11).

\begin{algorithm}
\begin{algorithmic}[1]
\Require{$\beta$, the number of activation scores to revise}
\Require{LibMR models $p_j$}
\Require{Activation vector $\bm{a} \in \mathcal{R}^{K \times 1}$ }
\State{Obtain sorted indices $\bm{b} = \textrm{argsort}(\bm{a})$}
\For{$i = 0,\dots,K-1$}{ $m_i = 1$}
\EndFor
\For{$i = 1,\ldots,\beta$}
\If{$a_{b_i} > a_j \forall j \neq i$}
   continue
\Else
  \State{$m_{\bm{b}_i} = 1 - \frac{\beta -
  i}{\beta}e^{-(\frac{\bm{a}_{\bm{b}_i}}{\lambda_{\bm{b}_i}})^{\kappa_{\bm{b}_i}}}$}
\EndIf
\EndFor
\State{Revise activation vector $\hat{\bm{a}} = \bm{a} \cdot \bm{m}$}
\State{Define $a_{K} = \sum_i (a_i - a_i \cdot m_i)$}
\State{$\hat{P}(y = j\,|\,\bm{x}) = \frac{e^{a_j}}{\sum\limits_{i =
0}^{K}e^{a_i}}$}
\State{Let $\hat{y} = \argmax_{j} \hat{P}(y = j\,|\,\bm{x})$}
\If{$\hat{y}$ is $K$} 
   reject input
\EndIf
\end{algorithmic}
\caption{MetaMax probability estimation using non-match activations.}
\label{alg:meta_estimation}
\end{algorithm}

\section{Experiments}
\label{sec:experiments}
In this section, we present the experimental details for testing MetaMax,
compare it against OpenMax and other recent work, and report the measured
performance on several public datasets. Additionally, to demonstrate the
applicability of MetaMax to enable closed-set classifiers to perform OSR,
we apply and report the performance of MetaMax to the following popular
classification networks: DenseNet \cite{huang2017densely}, ResNet
\cite{he2016deep}, and VGGNet \cite{simonyan2014very}.

\subsection{Experimental Setup}
Following experimental setup in \cite{yoshihashi2019classification}, we use
DenseNet \cite{huang2017densely} as the backbone network for classification.
We evaluate our method on the following common benchmark datasets: MNIST
\cite{lecun2010mnist}, SVHN \cite{netzer2011reading}, CIFAR10
\cite{krizhevsky2009learning}, and TinyImageNet (TIN) \cite{le2015tiny}.  To
test under open-set conditions, we follow the most common data partition
protocol \cite{neal2018open}. Specifically, for MNIST, SVHN, and CIFAR10, we
split each dataset at random such that 6 classes are chosen to be known and the
remaining 4 classes to be unknown. For the TinyImageNet dataset, we train on
$K=20$ known classes and test on the full 200-class set. We repeat the
experiment over 5 runs and report the average score. For the evaluation
metrics, we use the area under receiver operating characteristic (AUROC) curve
and the F1-scores.

\subsection{Results}
In the first set of experiments, we demonstrate the effectiveness of MetaMax.
We compare the AUROC score of MetaMax against recent methods in this area.  We
also show the multiclass ROC curves of our method on each dataset in the
appendix. The ROC curves demonstrate the magnitude of difficulty with each
dataset.

As shown in Table~\ref{tab:comparison}, MetaMax outperforms most recent methods
and is on par with the state of the art.  Note that most of these OSR methods
either need to \textit{modify the network architecture} or require
\textit{auxiliary training data}, which can be \textit{nontrivial} to obtain in
practice.  On the contrary, MetaMax is \textit{much simpler} and it can be
\textit{directly applied} post-hoc to a classifier with little-to-no extra
overhead.

In the second set of experiments, to verify the wide applicability of allowing
various classification networks to perform OSR using MetaMax, we apply and
report the F1-scores for two other networks: VGGNet \cite{simonyan2014very} and
ResNet \cite{he2016deep}. The results can be found in the appendix, which shows
that MetaMax outperforms OpenMax and the baseline network consistently. This
demonstrates the significance of our work in that MetaMax can potentially be
applied to any classification network and enable it to operate under open-set
conditions.

\begin{table}
\centering
\setlength{\abovecaptionskip}{-0.15cm}
\begin{center}
\begin{tabular}{|c|c|c|c|c|}
\hline
{\bf Method} & {\bf MNIST} & {\bf SVHN} & {\bf CIFAR10} & {\bf TIN} \\
\hline
\hline
G-OpenMax \cite{ge2017generative} & 0.984  & 0.896 & 0.675 & 0.580\\
OSRCI \cite{neal2018open} & 0.988  & 0.910 & 0.699 & 0.586\\
CROSR \cite{yoshihashi2019classification} & 0.991  & 0.899 & 0.883 & 0.589\\
C2AE \cite{oza2019c2ae} & 0.989  & 0.922 & 0.895 & 0.748\\
GDFR \cite{perera2020generative} & -  & 0.935 & 0.807 & 0.608 \\
CGDL \cite{sun2020conditional} & 0.994  & 0.935 & 0.903 & 0.762\\
OpenHybrid \cite{zhang2020hybrid} & 0.995  & 0.947 & 0.950 & 0.793\\
PROSR \cite{zhou2021learning} & -  & 0.943 & 0.891 & 0.693\\
OpenGAN \cite{kong2021opengan} & \bf{0.999} & \bf{0.988} & \bf{0.973} & \bf{0.907}\\
\hline
MetaMax (Ours) & 0.997 & 0.977 & 0.938 & 0.846 \\
\hline
\end{tabular}
\end{center}
\caption{AUROC scores of MetaMax and the compared methods using DenseNet. TIN
stands for TinyImageNet.}
\label{tab:comparison}
\end{table}

\section{Conclusion}
\label{sec:conclusion}
In this work we introduced MetaMax, an approach to calibrate deep neural
network-based classifiers by modeling non-match class activations for OSR.
Experiments on four standard image datasets demonstrate the effectiveness of
the proposed method. As a simpler and more effective alternative to OpenMax,
the modularization and general applicability of our method can have a wide
impact in the community and benefit future research by applying MetaMax to
calibrate standard closed-set classification networks for open-set conditions.

{\small
\bibliographystyle{ieee_fullname}
\bibliography{improved_open-set_deep_neural_networks_via_weibull_calibration}
}

\appendix

\section*{Appendix}
In this appendix, we provide additional experimental details on MetaMax.
First, we describe the training procedure and show results that highlight the
sensitivity of MetaMax to different hyperparameters.  Next, we illustrate the
performance differences among SoftMax, OpenMax, and MetaMax by analyzing their
respective ROC curves. We also use scatter plots to illustrate the correlation
between class activations and mean distances. Finally, we demonstrate the
applicability of MetaMax by providing comparisons against SoftMax and OpenMax
on DenseNet \cite{huang2017densely}, ResNet \cite{he2016deep}, and VGGNet
\cite{simonyan2014very}.

\section{Training Parameters}
In the first set of experiments where we showed the superiority of MetaMax over
OpenMax, we trained DenseNet121 for 200 epochs with a batch size of 64 for all
three datasets. DenseNet121 has a total of 6,960,006 parameters in its 121
layers. Training was performed using the Adam \cite{kingma2014adam}
optimization method. The learning rate decayed by a multiplicative factor
$\gamma$ at the end of each epoch. In the second set of experiments where we
demonstrated the applicability of MetaMax on different classification networks,
we conducted experiments using ResNet and VGGNet following the same
experimental protocol.

\section{Sensitivity to Hyperparameters}
The hyperparameter $q$ in Algorithm 1 is passed to the FitHigh function along
with the obtained non-match scores. $q$ controls the number of top non-match
activations we choose to fit the per-class Weibull model during the inference
step. We performed an experiment to show the sensitivity of MetaMax to this
hyperparameter. To do this, we varied the range of $q$ from 2 to 30. As shown
in Table~\ref{tab:auroc_scores_q}, the results are very robust to these
changes.

\begin{table}
\begin{center}
\begin{tabular}{|c|c|c|c|}
\hline
$q$ & {\bf F1-Score $\uparrow $} & {\bf AUROC $\uparrow $} \\
\hline
\hline
2  & 0.71107668 & 0.93882572 \\
\hline
5  & 0.71114378 & 0.93882158 \\
\hline
10  & 0.71109837 & 0.93880752 \\
\hline
20  & 0.71141039 & 0.93878481 \\
\hline
30  & 0.71089159 & 0.93872475 \\
\hline
\end{tabular}
\end{center}
\caption{The F1 and AUROC scores of MetaMax using DenseNet on CIFAR-10 with
different values of the hyperparameter $q$.}
\label{tab:auroc_scores_q}
\end{table}

\section{ROC Curves}
\begin{figure*}
\centering
\subfloat[]{\includegraphics[width=.52\columnwidth]{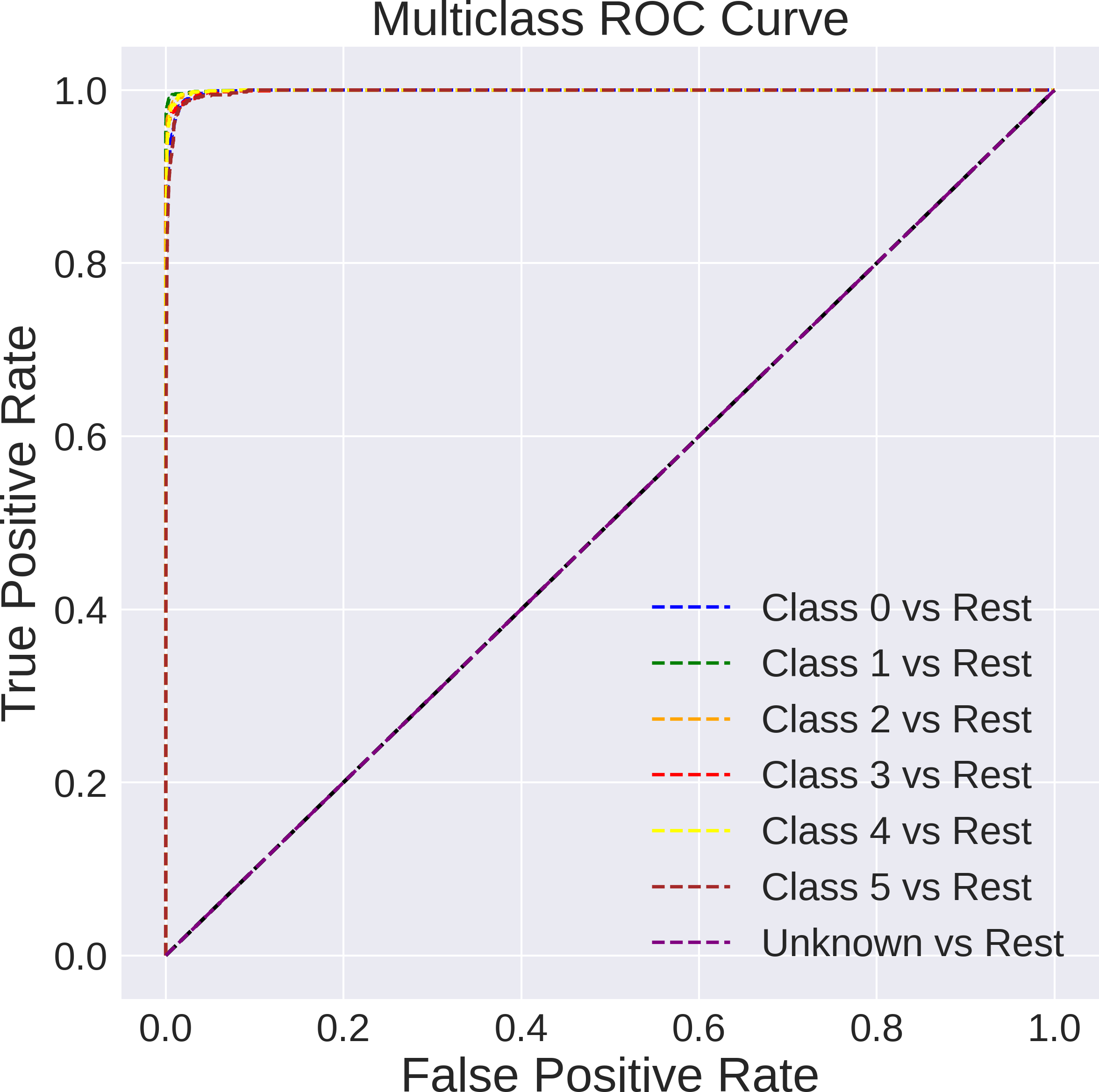}\hspace{1mm}}
\subfloat[]{\includegraphics[width=.52\columnwidth]{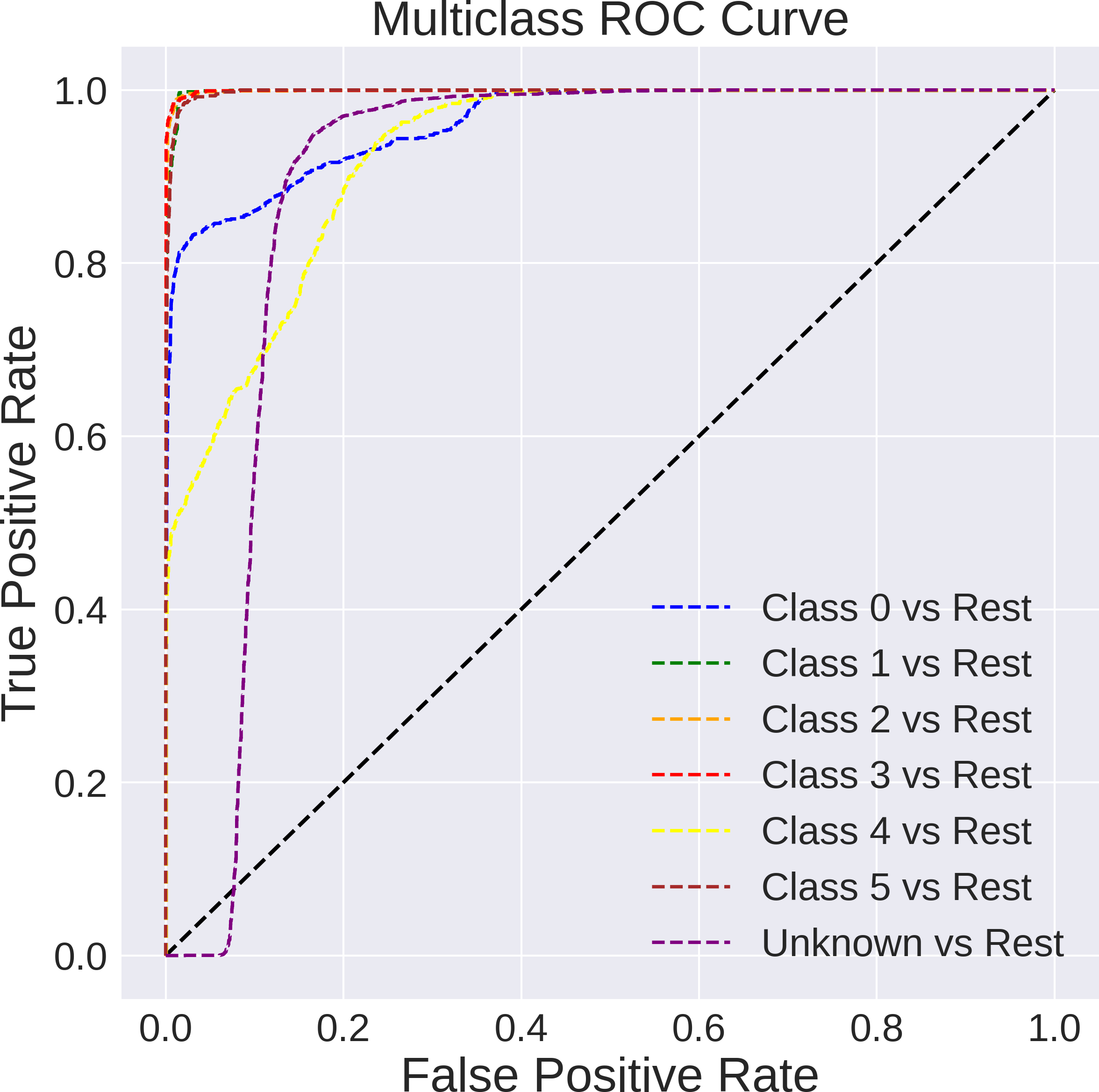}\hspace{1mm}}
\subfloat[]{\includegraphics[width=.52\columnwidth]{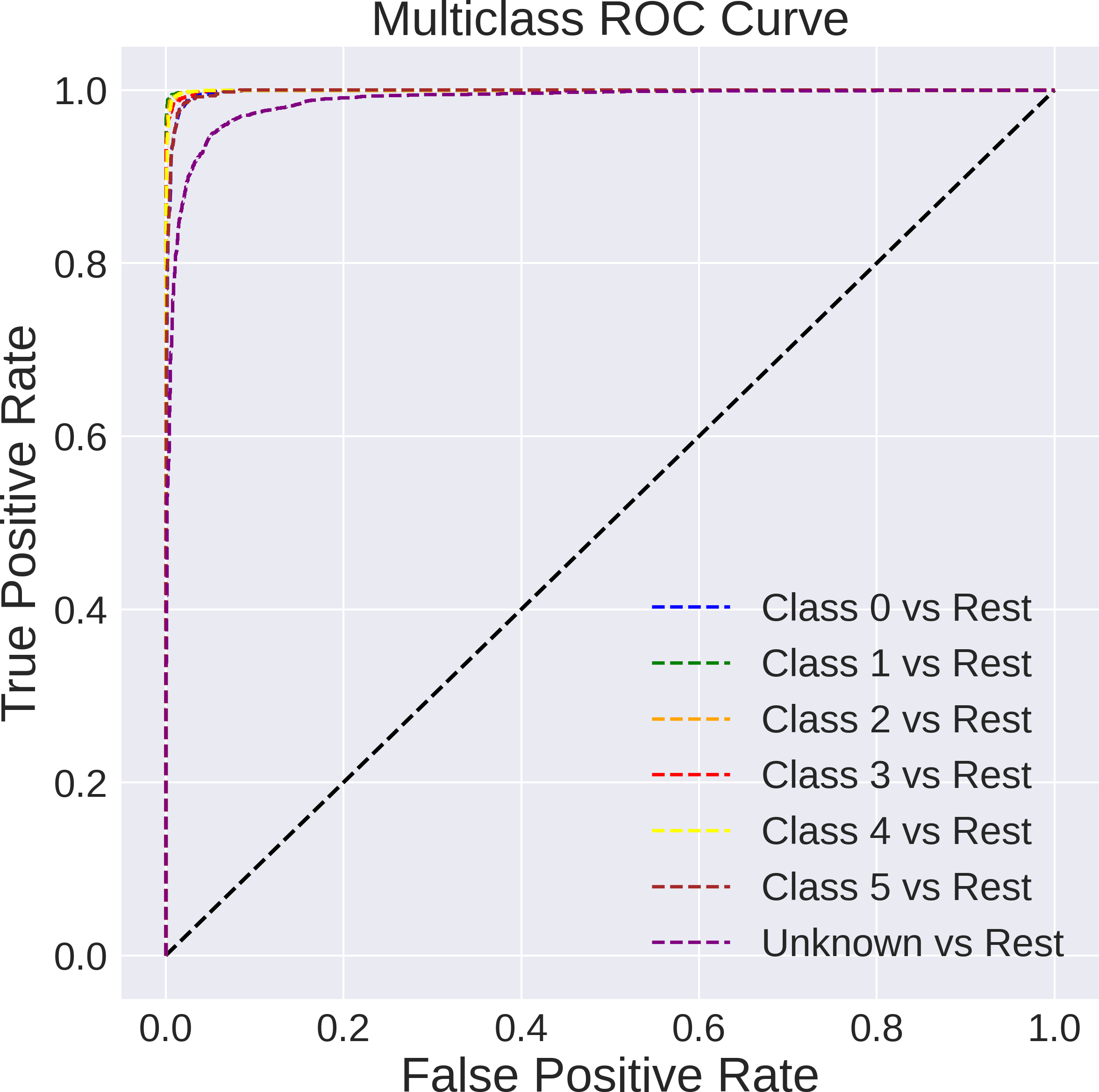}\hspace{1mm}}\\
\vspace{-2mm}

\subfloat[]{\includegraphics[width=.52\columnwidth]{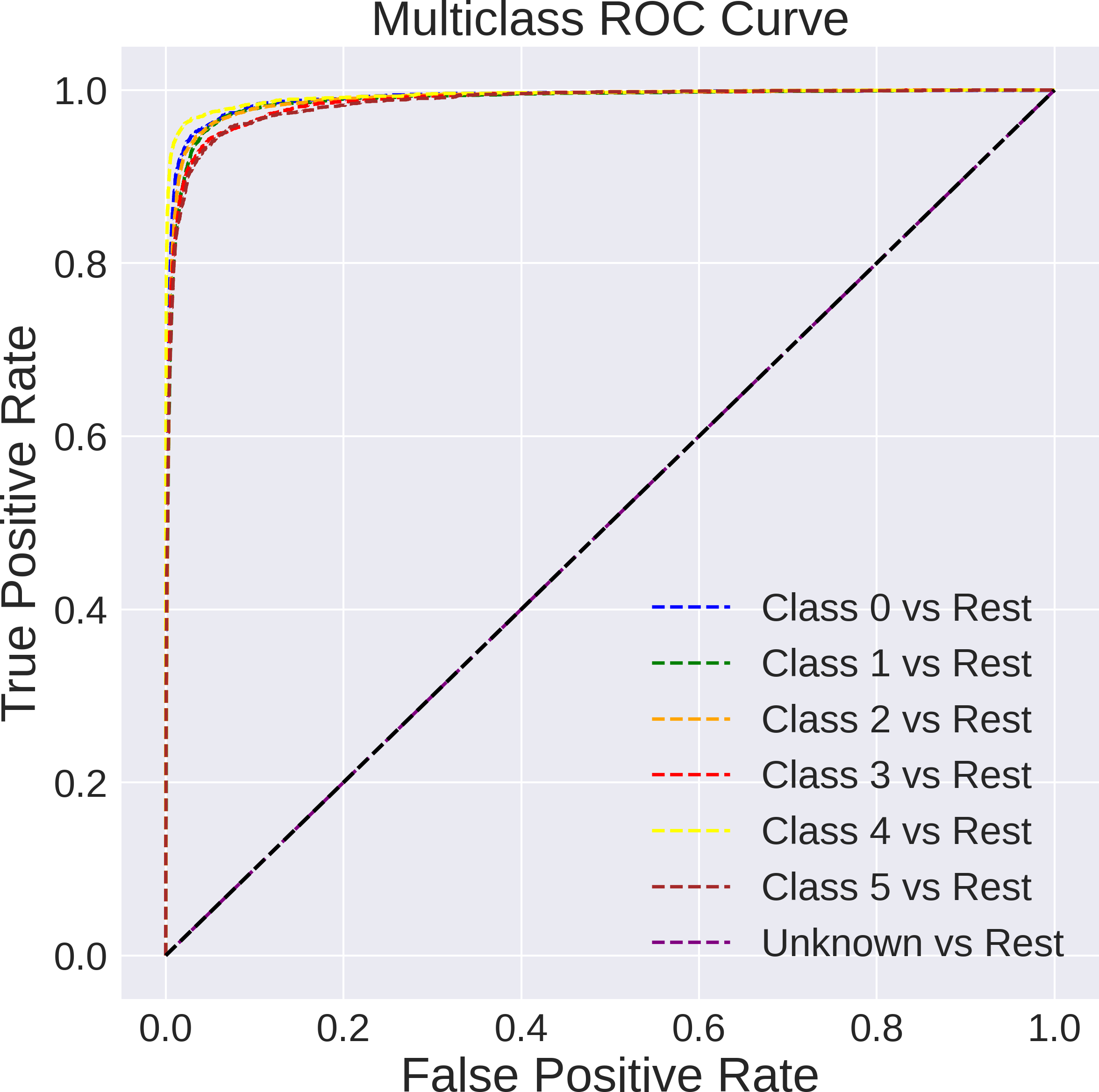}\hspace{1mm}}
\subfloat[]{\includegraphics[width=.52\columnwidth]{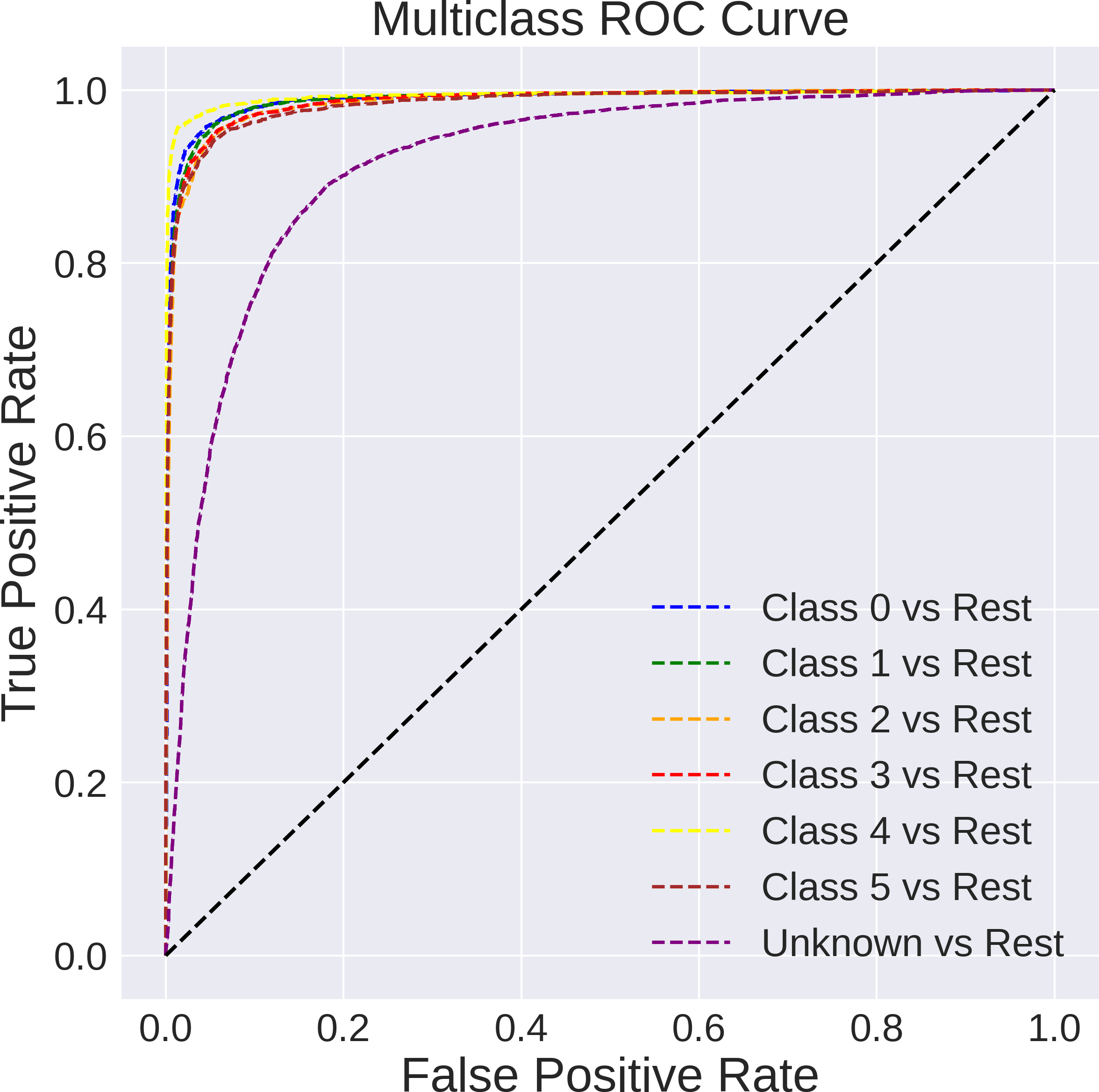}\hspace{1mm}}
\subfloat[]{\includegraphics[width=.52\columnwidth]{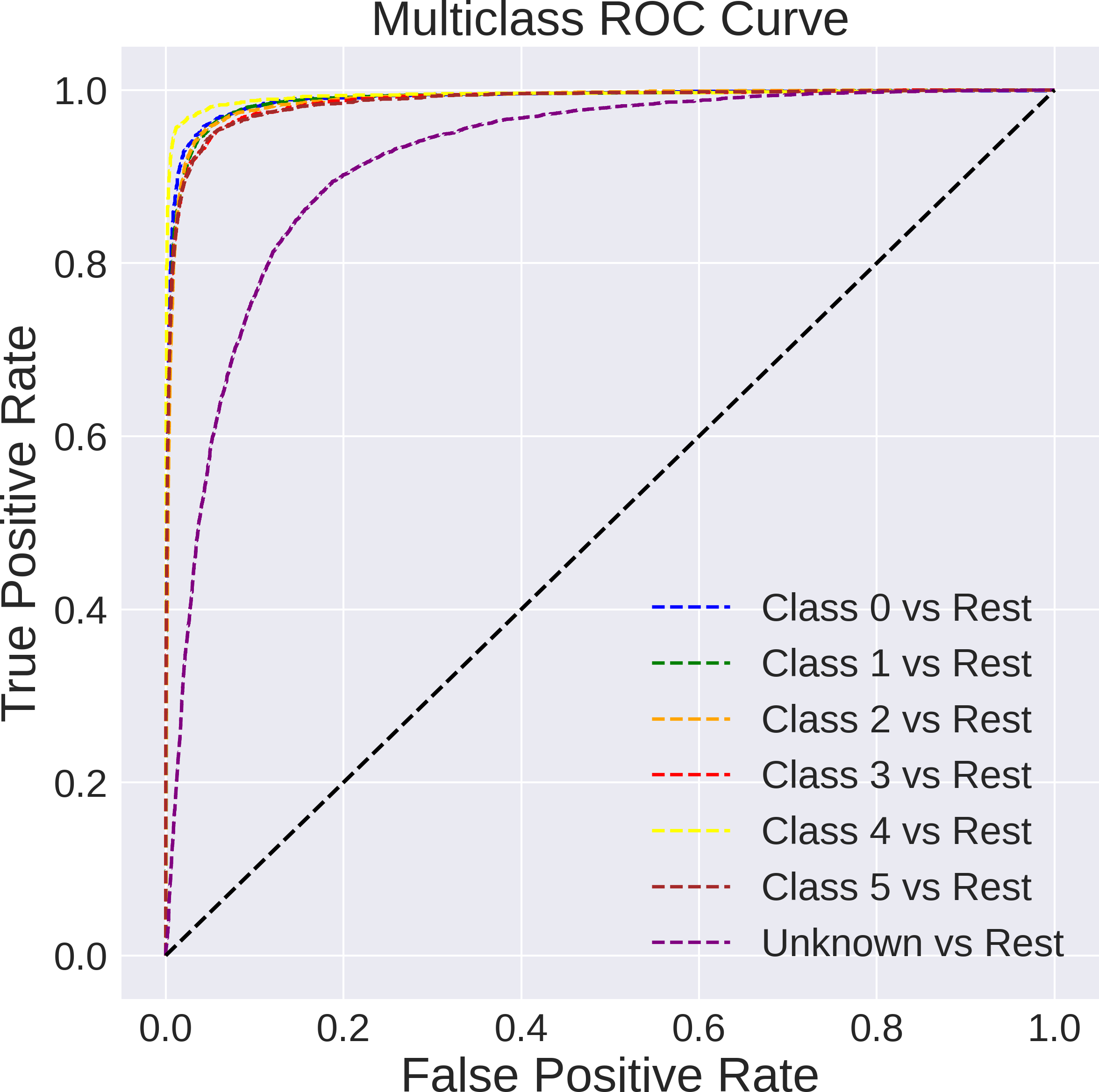}\hspace{1mm}}\\
\vspace{-2mm}

\subfloat[]{\includegraphics[width=.52\columnwidth]{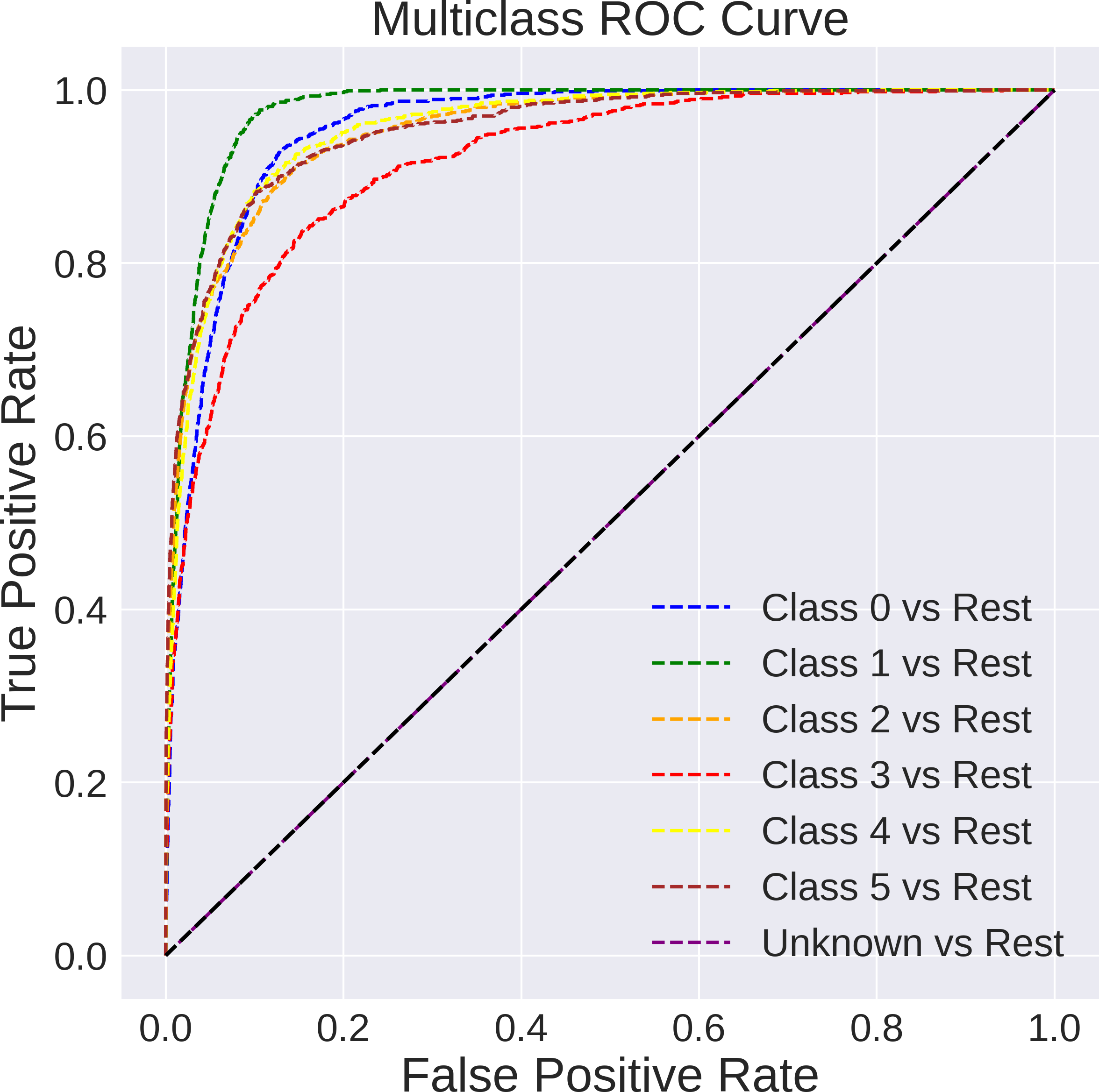}\hspace{1mm}}
\subfloat[]{\includegraphics[width=.52\columnwidth]{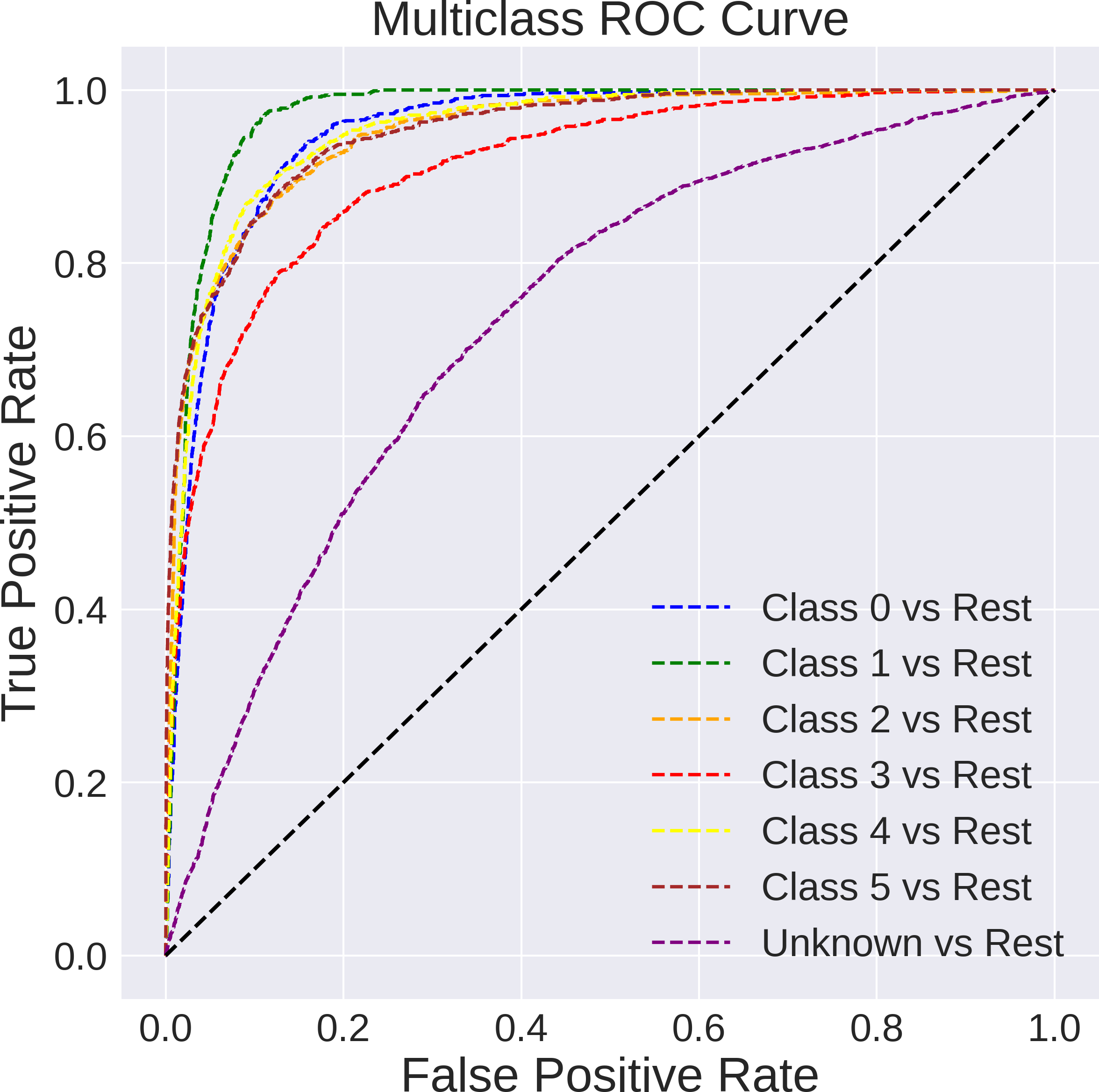}\hspace{1mm}}
\subfloat[]{\includegraphics[width=.52\columnwidth]{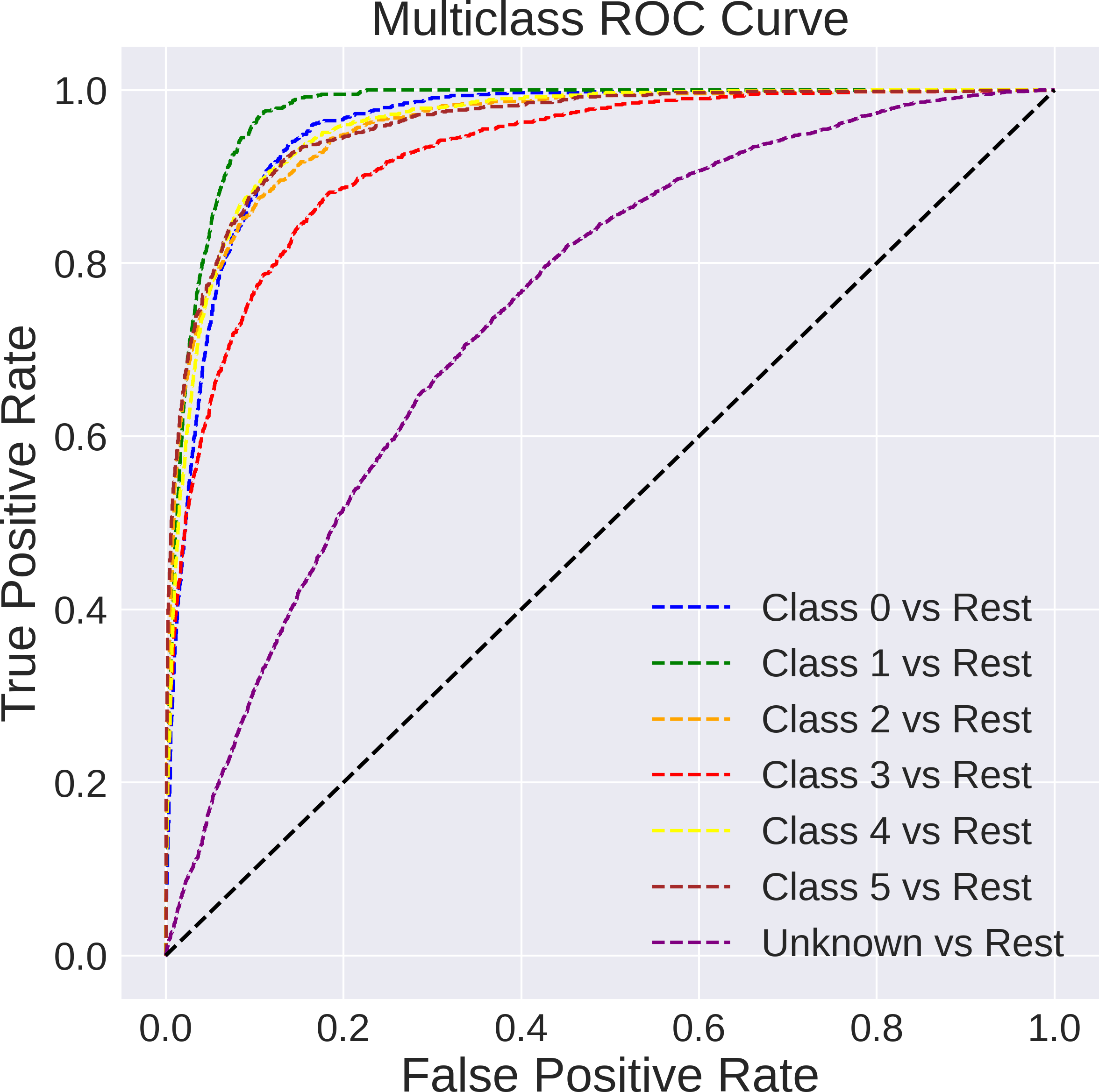}\hspace{1mm}}\\
\vspace{-10mm}

\subfloat[]{\includegraphics[width=.75\columnwidth]{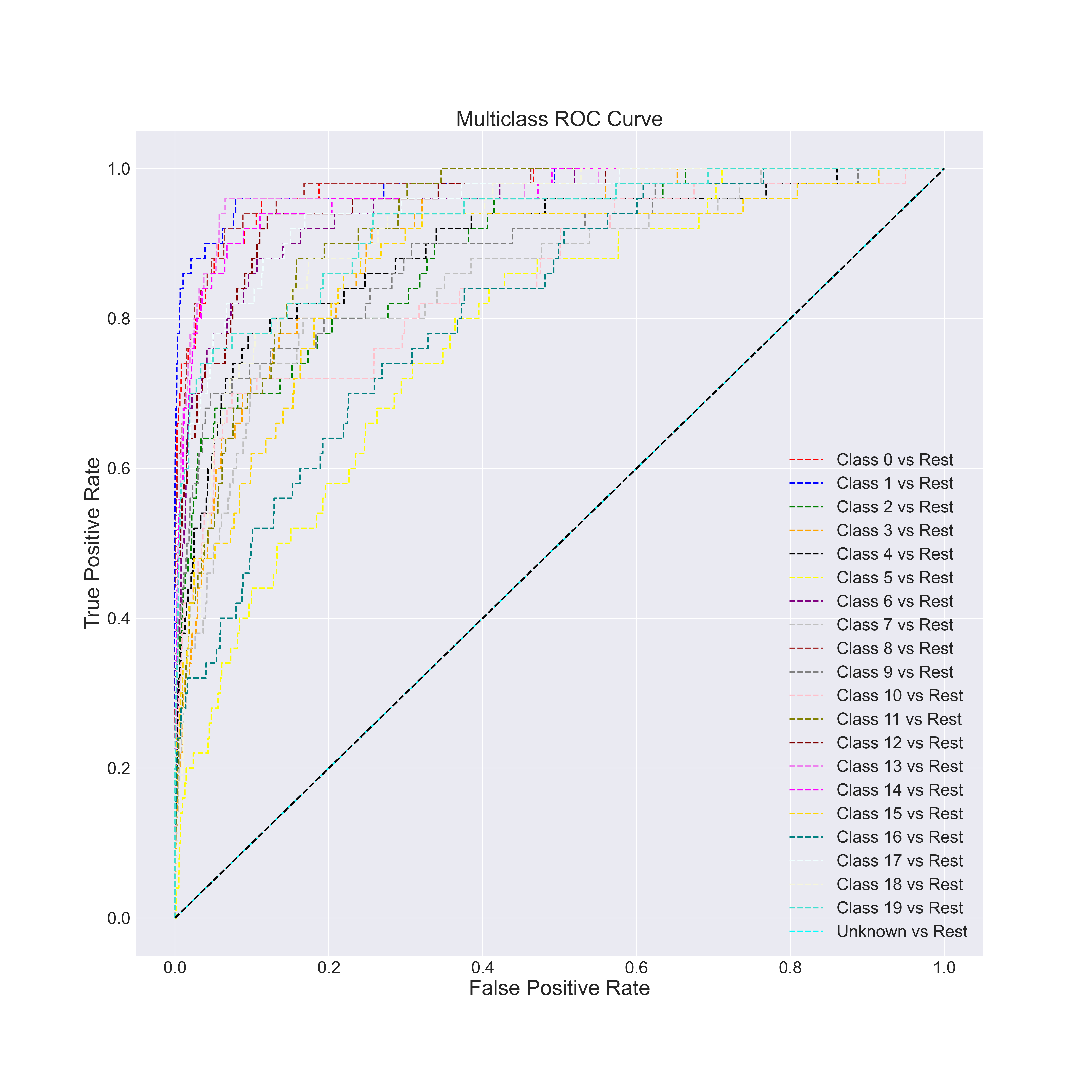}\hspace{-4mm}}
\subfloat[]{\includegraphics[width=.75\columnwidth]{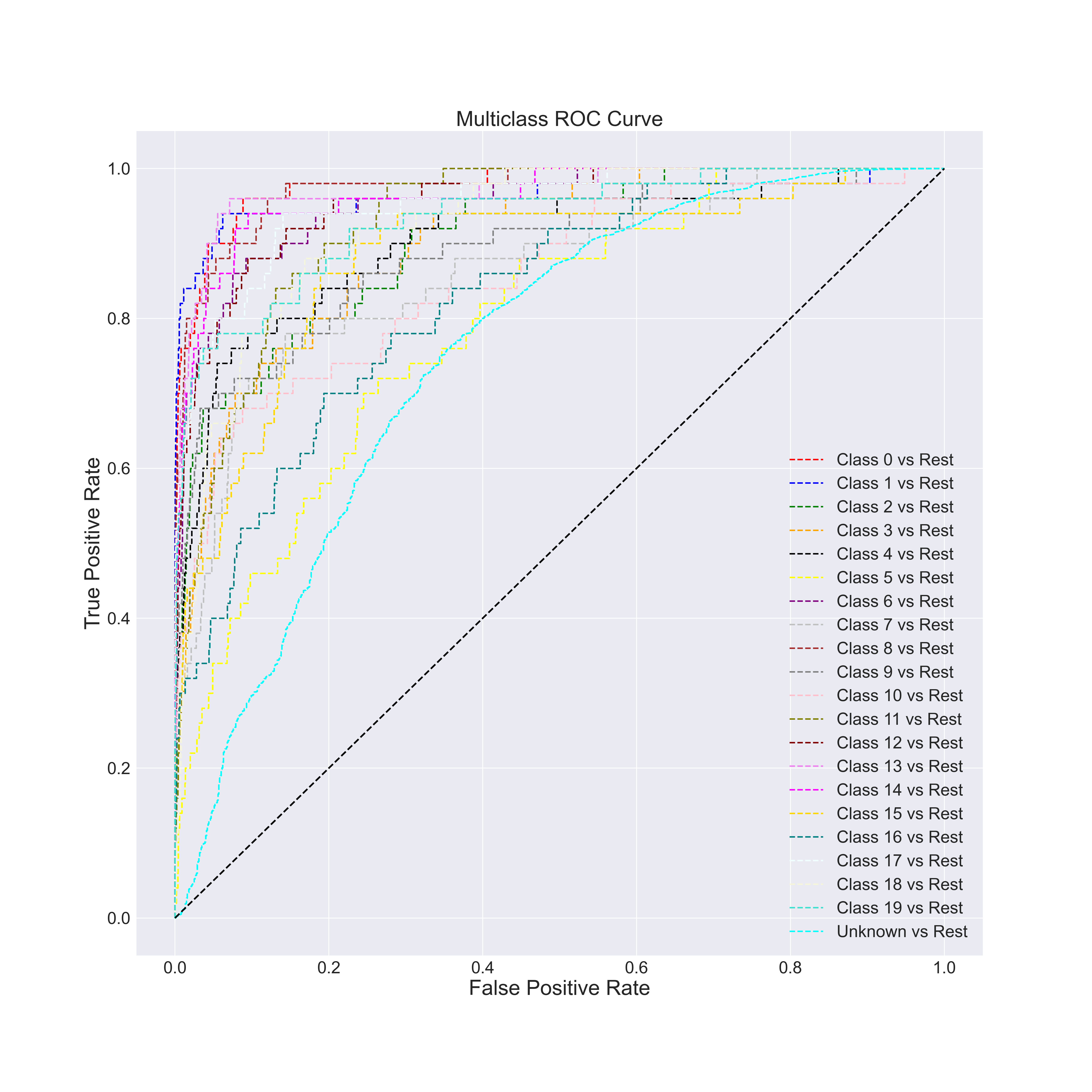}\hspace{-4mm}}
\subfloat[]{\includegraphics[width=.75\columnwidth]{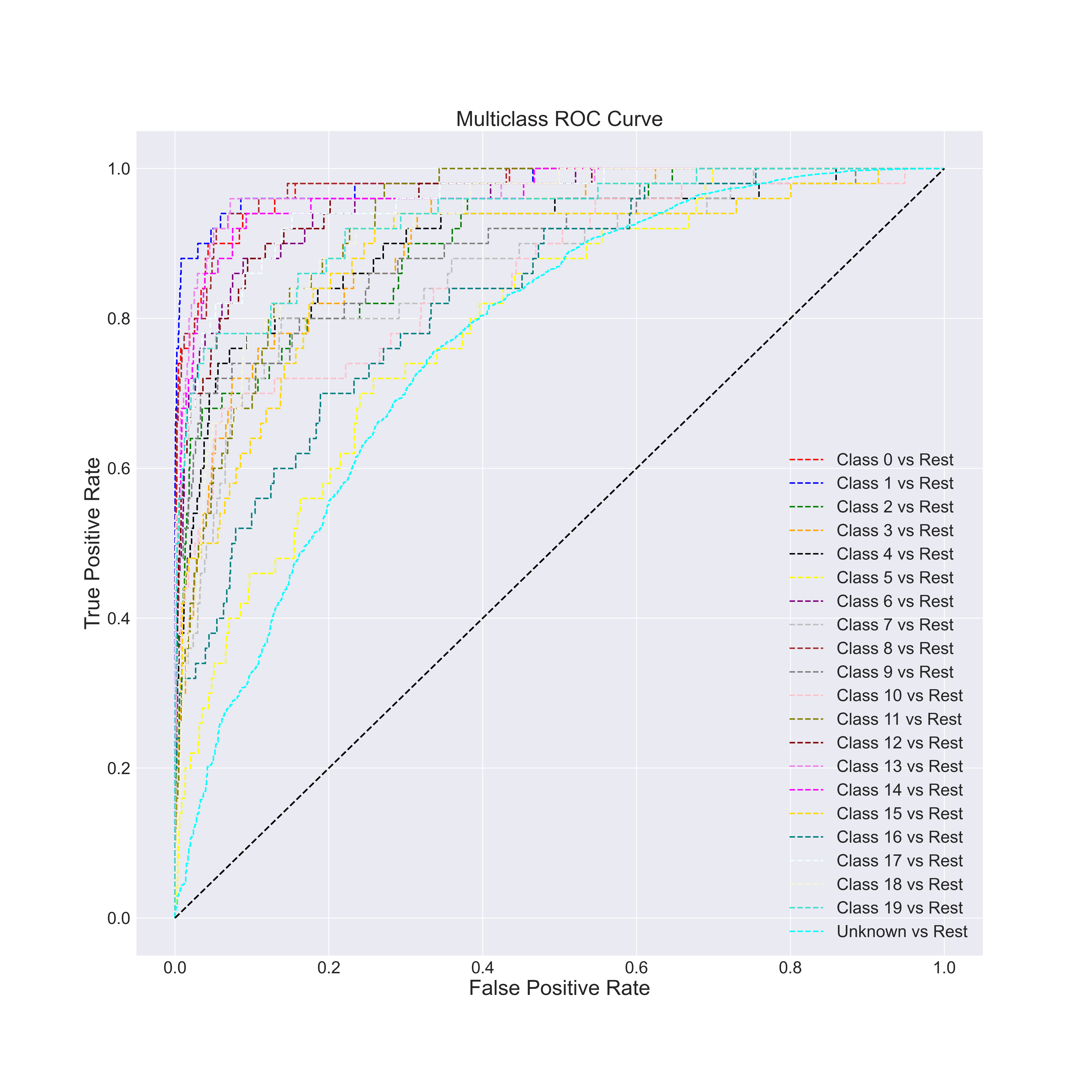}\hspace{0mm}}\\
\caption{The multiclass ROC curves for the (a-c) MNIST, (d-f) SVHN, (g-i) CIFAR10,
and (j-l) TinyImageNet datasets. The left, middle, and right columns illustrate the ROC
curves for SoftMax, OpenMax, and MetaMax, respectively.}
\label{fig:supp_roc_curves}
\end{figure*}

ROC curves demonstrating the magnitude of difficulty SoftMax, OpenMax, and
MetaMax have with each dataset are shown in Fig.~\ref{fig:supp_roc_curves}. In
the first column of Fig.~\ref{fig:supp_roc_curves}, the SoftMax activation
function has steeper ROC curves for all classes except the unknown class.
However, SoftMax's unknown ROC curve lies directly on top of the
non-discrimination line, which dramatically hurts performance. OpenMax and
MetaMax both have steep ROC curves for the unknown class. Nevertheless, MetaMax
has a steeper curve when compared against OpenMax. In comparison to SoftMax and
OpenMax, this aligns with the higher AUROC scores for MetaMax as shown in the
main paper. The area under each class's ROC curve can be interpreted as being
inversely related to the difficulty of picking a sample from that class against
all other classes.

\section{Scatter Plots}
\begin{figure*}
\centering
\setlength{\abovecaptionskip}{0.11cm}
\subfloat[]{
    \includegraphics[width=.9\columnwidth]{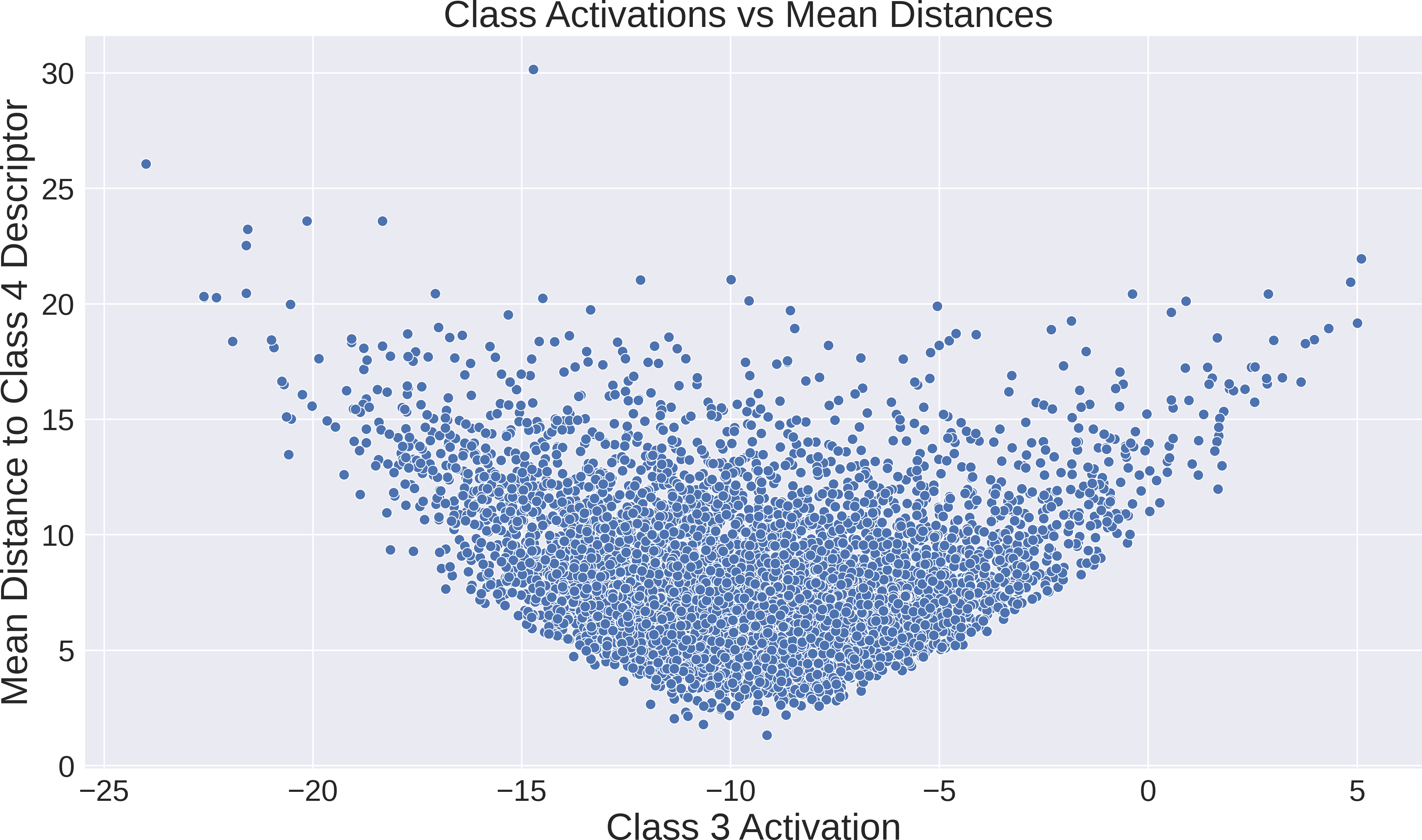}\hspace{1mm}
    \label{subfig:mean_vs_activations_svhn}}
\subfloat[]{
    \includegraphics[width=.9\columnwidth]{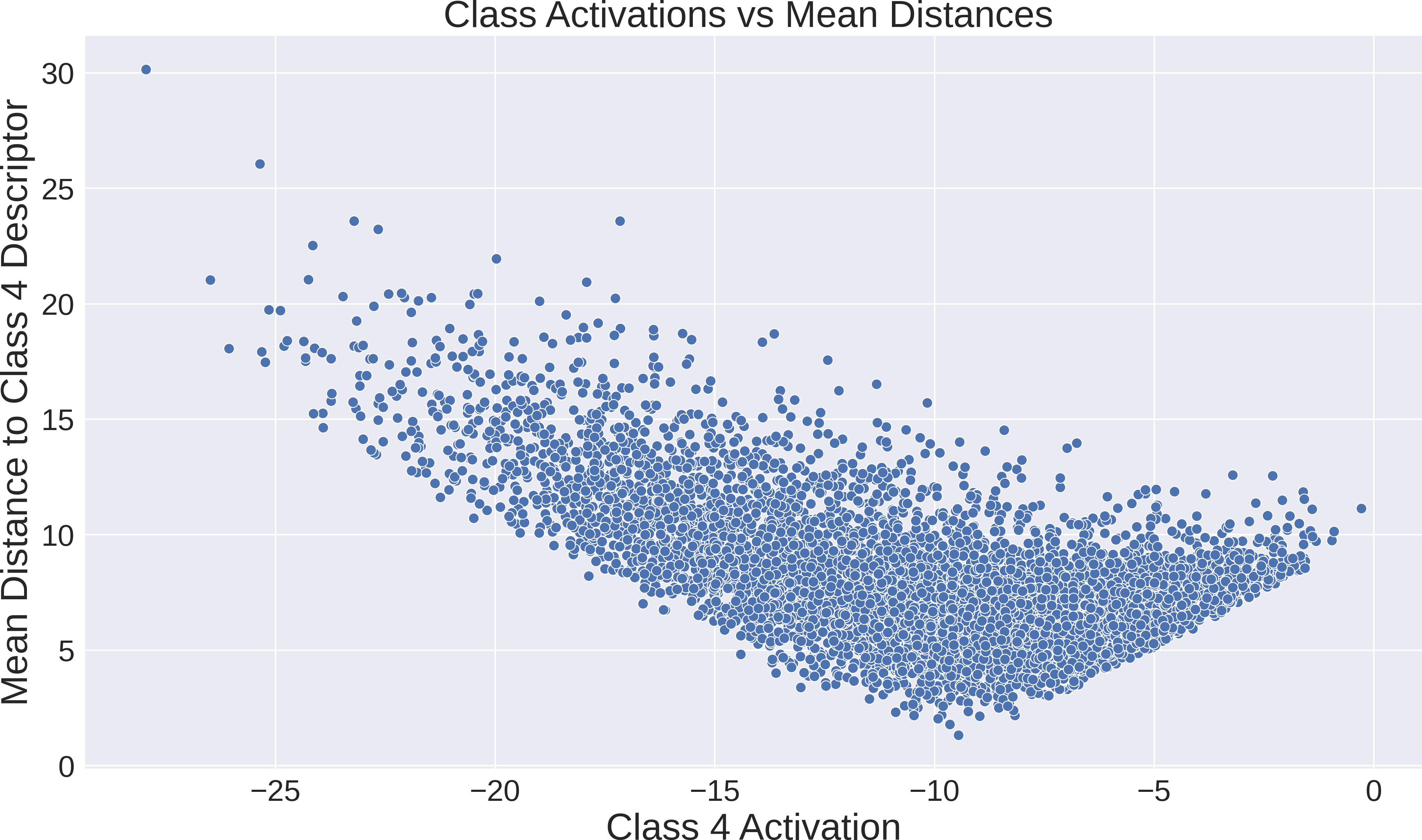}\hspace{1mm}
    \label{subfig:mean_vs_activations_svhn_match}}\\
\caption{An illustration of the correlation between class activations and mean
distances for (a) non-match and (b) match activation vectors. Both plots
correspond to the activation vectors associated with class 4 in the SVHN
\cite{netzer2011reading} dataset. Figures (a) and (b) show plots of the
activations of class 3 and 4, respectively, on the x-axis.}
\label{fig:mean_vs_activations_svhn_match}
\end{figure*}

Intuitively, class activations offer a measure of similarity between a sample
and its implicitly-stored mean activation vectors in the latent space. We
present empirical evidence of this in
Fig.~\ref{fig:mean_vs_activations_svhn_match} using the SVHN dataset.
Fig.~\ref{subfig:mean_vs_activations_svhn} shows a  scatter plot for the
activation vectors of class 4. The distance of each activation vector to the
mean activation for class 4 is plotted on the y-axis. On the x-axis are the
class activations for class 3. We can see that as the non-match activation
increases, the distance from this sample to the class 4 descriptor also
increases. We believe this correlation justifies the use of implicit over
explicit class descriptors. Furthermore, it suggests that activations are
highly correlated with the similarity between a sample and a class descriptor.
In Fig.~\ref{subfig:mean_vs_activations_svhn_match} we show a plot of class-4
class activations versus mean distances to the class-4 descriptor.

\section{ResNet and VGGNet Results}
To verify the wide applicability of various classification networks in
performing open-set recognition using MetaMax, we utilize and report the
F1-scores on DenseNet (Table~\ref{tab:f1_scores}), ResNet
(Table~\ref{tab:resnet_f1_scores}), and VGGNet (Table~\ref{tab:vgg_f1_scores}).
We can see from these results that MetaMax consistently outperforms both OpenMax
and the baseline network. This demonstrates the significance of our work in that
MetaMax can be applied to any classification network, hence allowing it to
operate under open-set conditions. Moreover, these results provide evidence that
other open-set recognition methods can achieve performance gains using MetaMax
alongside any backbone network.

\begin{table}
\footnotesize{
\begin{center}
\begin{tabular}{|c|c|c|c|c|}
\hline
{\bf Method} & {\bf MNIST} & {\bf SVHN} & {\bf CIFAR10} & {\bf TinyImageNet}\\
\hline
\hline
SoftMax & 0.644 & 0.682 & 0.547 & 0.512\\
OpenMax & 0.758 & 0.815 & 0.669 & 0.627\\
\hline
MetaMax (Ours) & \bf{0.813} & \bf{0.846} & \bf{0.711} & {\bf 0.683}\\
\hline
\end{tabular}
\end{center}
\caption{The F1-scores of MetaMax against other methods using DenseNet.}
\label{tab:f1_scores}
}
\end{table}

\begin{table}
\footnotesize{
\begin{center}
\begin{tabular}{|c|c|c|c|c|}
\hline
{\bf Method} & {\bf MNIST} & {\bf SVHN} & {\bf CIFAR10} & {\bf TinyImageNet} \\
\hline
\hline
SoftMax & 0.637 & 0.684 & 0.542 & 0.510\\
OpenMax & 0.735 & 0.828 & 0.654 & 0.618\\
\hline
MetaMax (Ours) & \bf{0.801} & \bf{0.854} & \bf{0.696} & {\bf 0.669}\\
\hline
\end{tabular}
\end{center}
\caption{The F1-scores of MetaMax against other methods using ResNet.}
\label{tab:resnet_f1_scores}
}
\end{table}

\begin{table}
\footnotesize{
\begin{center}
\begin{tabular}{|c|c|c|c|c|}
\hline
{\bf Method} & {\bf MNIST} & {\bf SVHN} & {\bf CIFAR10} & {\bf TinyImageNet}\\
\hline
\hline
SoftMax & 0.623 & 0.671 & 0.528 & 0.496\\
OpenMax & 0.695 & 0.793 & 0.607 & 0.583\\
\hline
MetaMax (Ours) & \bf{0.752} & \bf{0.817} & \bf{0.654} & \bf{0.618} \\
\hline
\end{tabular}
\end{center}
\caption{The F1-scores of MetaMax against other methods using VGGNet.}
\label{tab:vgg_f1_scores}
}
\end{table}

\end{document}